\documentclass{article}
\usepackage{spconf,amsmath,graphicx}
\usepackage{hyperref}
\usepackage{comment}

\usepackage{enumitem}
\setlist{nosep, leftmargin=14pt}

\usepackage{mwe} % to get dummy images

\title{A Multi-Stage Optimization Pipeline for Bethesda Cell Detection in Pap Smear Cytology}
\name{Martín Amster$^{1,2}$ \qquad Camila María Polotto$^{2,3}$ \thanks{© 2026 IEEE.  Personal use of this material is permitted.  Permission from IEEE must be obtained for all other uses, in any current or future media, including reprinting/republishing this material for advertising or promotional purposes, creating new collective works, for resale or redistribution to servers or lists, or reuse of any copyrighted component of this work in other works.}}

\address{$^{1}$ Instituto de Ciencias de la Computación (ICC), CONICET-UBA \\
         $^{2}$ Facultad de Ciencias Exactas y Naturales, Universidad de Buenos Aires \\
         $^{3}$ Instituto de Física Interdisciplinaria y Aplicada, Departamento de Física, Universidad de Buenos Aires}
\begin{document}

%\ninept
%
\maketitle

\begin{abstract}
Computer vision techniques have advanced significantly in recent years, finding diverse and impactful applications within the medical field. In this paper, we introduce a new framework for the detection of Bethesda cells in Pap smear images, developed for Track B of the Riva Cytology Challenge held in association with the International Symposium on Biomedical Imaging (ISBI). This work focuses on enhancing computer vision models for cell detection, with performance evaluated using the mAP50-95 metric. We propose a solution based on an ensemble of YOLO and U-Net architectures, followed by a refinement stage utilizing overlap removal techniques and a binary classifier. Our framework achieved second place with a mAP50-95 score of 0.5909 in the competition. The implementation and source code are available at the following repository: \href{https://github.com/martinamster/riva-trackb}{github.com/martinamster/riva-trackb}.
\end{abstract}

\section{Introduction}
\label{sec:intro}

Cervical cancer is the fourth most prevalent cancer and the fourth leading cause of cancer death in women, with an estimated 604,000 new cases and 342,000 deaths worldwide in 2020 \cite{Sung_2021}. It remains a critical global health priority, with great importance placed on early intervention. While the widespread implementation of Pap smear screening has fundamentally transformed early detection and patient outcomes \cite{Perkins_2023}, the efficacy of these programs still relies heavily on the manual interpretation of cytological samples. This traditional approach is inherently labor-intensive and susceptible to human factors, including inter-observer variability and fatigue under high-volume workloads, which can compromise diagnostic sensitivity. %Falta cita?

\begin{figure}[t]

\begin{minipage}[b]{1.0\linewidth}
  \centering
  \centerline{\includegraphics[width=7.5cm]{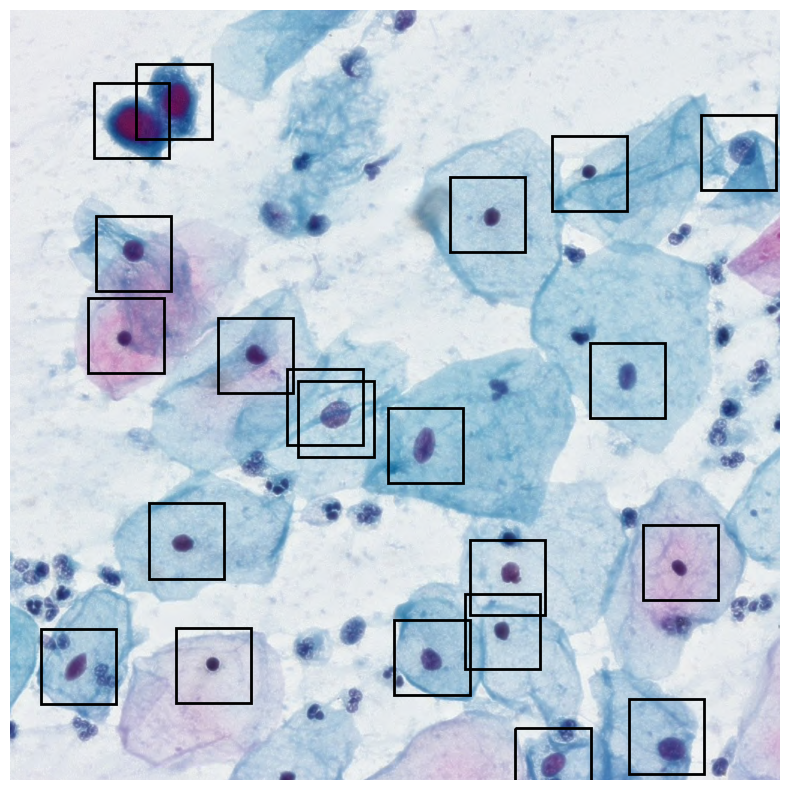}}
%  \vspace{2.0cm}
 % \centerline{(a) Result 1}\medskip
\end{minipage}
\caption{Example of an annotated image with a 100×100 bounding box from the RIVA dataset.}
\label{fig:annotated_cells}
\end{figure}

The integration of deep learning offers a transformative path to support cytologists by automating the classification of healthy versus pathological cells \cite{Zhou_2021}. However, despite significant progress in computational cell analysis% (e.g.,  \cite{Amitay_2023, Alahmari_2024, Shifat_2020, Thomas_2017, Bhattarai_2024, Liu_2019, Shrestha_2023})%
, achieving robust generalization in Pap smear classification remains a persistent challenge. A primary obstacle is the lack of large-scale, high-quality datasets that reflect the biological diversity encountered in clinical practice \cite{Matias_2021}. To bridge this gap, Perez Bianchi et al. (2025) \cite{Perez_2025_RIVA} introduced the RIVA dataset, providing a standardized benchmark for the development of advanced detection and classification algorithms. Figure \ref{fig:annotated_cells} shows an example of an annotated image from the dataset.

Within the framework of the RIVA Cytology Challenge, Track B focuses specifically on the localization of these cells using bounding boxes. The goal is to evaluate the precision of detection models in complex cytological environments, utilizing the mean Average Precision (mAP50-95) as the primary metric for performance assessment.

%En la Figura~\ref{fig:entropy}
%\begin{figure}[H]
%\begin{center}
% \includegraphics[width=0.9\textwidth]%{tesis_entropy.png}
 %\end{center}
 %\caption{Ejemplo de agrupamiento semántico de %respuestas generadas por un LLM}
 %\label{fig:entropy}
%\end{figure}

 % Hasta aca llega lo que sería "correcto", después hay que reescribir todo para que no nos acusen de plagio igual

\section{Methodology}
\label{sec:methods}

\subsection{Pre-processing and YOLO model training}
\label{ssec:subhead}
Prior to model training, we addressed the limitation of the dataset's uniform 100×100 pixel annotations, which frequently failed to represent actual cell morphology. To optimize center localization, we resized the annotations to fixed bounding box dimensions ranging from 10×10 to 120×120 pixels. This resizing was performed by preserving the original center coordinates of each box while systematically adjusting its width and height. During inference, the model's predicted width and height were standardized back to the fixed 100×100 format. Empirical benchmarks identified 50×50 and 20×20 as the most effective scales for detection in YOLO models.

We implemented the YOLOv8 architecture (Nano, Small, and Medium variants), selected for its  rapid training and experimentation capabilities, allowing us to iterate through multiple configurations efficiently. Additionally, the model consistently achieved high recall rates across all variants, which was critical for our detection task where missing cells is more costly than false positives. 

\subsection{Ensemble-based approach}
\label{ssec:subhead}
To improve detection precision, we implemented an ensemble strategy combining the outputs of two models. We defined a maximum Euclidean distance threshold in pixels between centroids to determine whether two predictions corresponded to the same cell. Detections identified by both models were retained, and their final center coordinates and confidence scores were calculated as the average of the two individual predictions. Conversely, detections predicted by a single model were only included in the final output if their confidence score exceeded an empirically determined threshold.

Our first ensemble integrated two YOLOv8n models, trained specifically with 20×20 and 50×50 pixel bounding boxes to capture varying levels of spatial context. Both YOLO models were trained for up to 150 epochs at an input resolution of 1024×1024 (batch size 16), and test-time augmentation was applied at inference to further boost recall. For this ensemble, we used a distance threshold of 12 pixels and set the empirical confidence threshold to 0.35.

To further refine the detection pipeline, a U-Net model with a ResNet34 backbone pre-trained on ImageNet was utilized. The model was trained as a heatmap regressor: for each annotated cell center, a 2D Gaussian kernel ($\sigma$ proportional to 1/6 of the expected box size at the target resolution) was rendered onto a float32 heatmap, and the network was optimized with MSE loss to reproduce these heatmaps. Training was conducted for 15 epochs with a batch size of 4, the Adam optimizer (lr = 0.0001), and images resized to 1024×1024, augmented with random horizontal and vertical flips. At inference, a multi-scale strategy was applied by averaging predictions at three resolution scales (0.8×, 1.0×, 1.2×), and cell centers were extracted from the resulting heatmap via local-maximum suppression using a 25×25 max-pooling kernel at a confidence threshold of 0.2. Although this segmentation-based approach exhibited lower recall than the YOLO detectors, it demonstrated significantly higher precision, providing complementary high-quality detections to the ensemble. We applied our previously described ensemble strategy a second time to merge the output of the first ensemble with the U-Net predictions, using a distance threshold of 12 pixels and a confidence threshold of 0. Consequently, for spatially coincident detections across both methods, the center coordinates and confidence scores were averaged. Furthermore, by setting the confidence threshold to zero, all unmatched detections originating exclusively from either the YOLO ensemble or the U-Net model were fully retained. This two-stage strategy effectively boosted the confidence of corroborated detections while preserving the complete set of individual predictions.

\subsection{Optimization pipeline}

\begin{figure*}
\begin{center}
    \includegraphics[width = 15.5 cm]{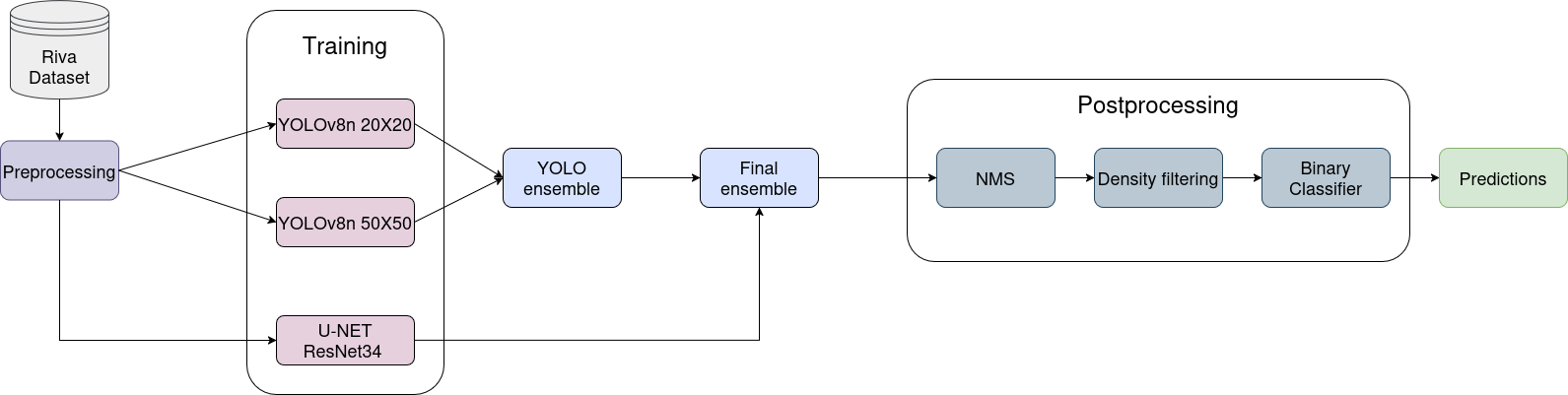}
\end{center}
 \caption{Overview of the architecture and optimization pipeline. Models are trained with preprocessed images to detect Bethesda cells. The YOLO ensemble is created using the output from both YOLO models. The final ensemble is created combining the YOLO ensemble with the output of the U-Net model. Final predictions are obtained from the three-step postprocessing pipeline.}
 \label{fig:pipeline}
\end{figure*}

\label{ssec:subhead}
The final ensemble's output was refined through a three-stage pipeline to optimize detection precision and remove background noise.

\textbf{Step 1: Non-Maximum Suppression (NMS).} After ensembling, duplicate detections arising from overlapping predictions across models were suppressed using NMS. Candidates were ranked by confidence and iteratively filtered: any box overlapping a higher-confidence detection by an IoU greater than 0.75 was discarded. This threshold was chosen to be deliberately permissive for two complementary reasons. First, it preserves nearby but distinct cells that are densely packed, avoiding the merging of legitimate adjacent detections. Second, the competition metric is mAP averaged over IoU thresholds from 0.50 to 0.95 (mAP50-95), which rewards predictions that tightly localise the target across a wide range of overlap levels. Suppressing only at 0.75 retains candidate boxes with slightly different positions, increasing the likelihood that at least one of them achieves a high IoU match with the ground truth at each evaluation threshold, thus contributing to higher average precision across the full metric range.

\begin{comment}
    
The $IoU$ is defined as:
\begin{equation}
IoU = \frac{\text{Area}(A \cap B)}{\text{Area}(A \cup B)}
\label{eq:iou}
\end{equation}
where $A$ and $B$ denote the areas of each predicted bounding box.
\end{comment}

\textbf{Step 2: Spatial Density Filtering.} Each image was partitioned into a 4×4 grid of equal quadrants. In high-density quadrants (at least 30 detections), the elevated volume of candidates often correlates with model uncertainty, which generates numerous low-confidence spurious predictions, so a confidence threshold of 0.1 was applied. In low-density quadrants (less than 30 detections), where even weak signals may correspond to isolated abnormal cells, the threshold was relaxed to 0.001 to avoid discarding potentially relevant detections.

\textbf{Step 3: Binary Classification Refinement.} To mitigate false positives in low-confidence ensemble predictions ($<$0.01), an EfficientNet-B0 binary classifier was trained to distinguish legitimate cellular morphology from background artifacts. The training set was constructed via hard negative mining: ground-truth boxes were labeled as cell, while predictions with an IoU $<$ 0.1 relative to any ground truth were labeled as garbage. The model was fine-tuned using a weighted binary cross-entropy loss to account for class imbalance and standard augmentations (flips, rotations, and color jitter). During inference, a binary threshold of 0.05 was applied to filter out spurious detections while preserving high sensitivity for pathological cases.\\

The final architecture and optimization pipeline can be found in Figure \ref{fig:pipeline}.

\begin{table*}
\begin{center}
    \begin{tabular}{|c|c|c|c|c|c|c|}                             \hline
    Model & TP & FP & Recall & Precision & F1 & mAP50-95\\                \hline
    YOLOv8n 20×20 & 2654 & 30472 & 0.9896 & 0.0801 & 0.1482 & 0.5966 \\ \hline
    YOLOv8n 50×50 & \textbf{2665} & 37398 & \textbf{0.9937} & 0.0665 & 0.1247 & 0.6059 \\ \hline
    YOLO ensemble & 2647 & 19403 & 0.987 & 0.12 & 0.2141 & 0.6153 \\ \hline
    U-Net & 2141 & \textbf{944} & 0.7983 & \textbf{0.694} & \textbf{0.7425} & 0.5494 \\ \hline
    \textbf{Final prediction} & 2544 & 11410 & 0.9485 & 0.1823 & 0.3058 & \textbf{0.6232} \\ \hline
    
\end{tabular}
\end{center}
 \caption{Quantitative evaluation on the RIVA validation set. The metrics are calculated using the models' output and ground-truths.}
 \label{fig:table}
\end{table*}

    \subsection{Evaluation metrics}
\label{ssec:subhead}
Model performance was evaluated using standard detection metrics (TP, FP, Recall, Precision,  F1) \cite{Rainio_2024} and the official competition metric, mAP50-95.
\begin{comment}
\textbf{TP:} number of predictions that match a ground-truth box (more than 0.5 IoU).

\textbf{FP:} number of predicted boxes without a matching ground-truth box.

\textbf{Recall:} 
\begin{equation*}
Rec. = \frac{\text{TP}}{\text{TP + FN}}
\end{equation*}
Where \textbf{FN} is the number of ground-truth boxes without a matching prediction.

\textbf{Precision:} 
\begin{equation*}
Pre. = \frac{\text{TP}}{\text{TP + FP}}
\end{equation*}

\textbf{F1:} 
\begin{equation*}
F1 = \frac{2*\text{Pre.}*\text{Rec.}}{\text{Pre. + Rec.}}
\end{equation*}

\textbf{mAP50-95:} mean value of the Average Precision (AP) calculated at IoU thresholds ranging from 0.5 to 0.95 with an increment of 0.05, where AP is defined as the area under the precision-recall curve, which is generated by plotting precision against recall at all possible confidence thresholds.
\end{comment}

All model and ensemble parameters were determined empirically to maximize the mAP50-95 on the test dataset. Because the ground-truth annotations for the test set were withheld by the organizers, performance on this split could only be evaluated by submitting our predictions to the official Kaggle challenge platform. Consequently, to provide a comprehensive and transparent analysis of our approach, the detailed performance metrics reported in the Results section are based on our internal validation dataset.

\section{Experimental Setup}
\label{sec:setup}
The study utilized a dataset of 959 high-resolution Pap smear patches (1024×1024 pixels), partitioned into 828 training and 131 validation images. Each image contains cells with standardized 100×100 pixel ground-truth bounding boxes centered on the cellular nuclei. All model training and inference were conducted within the Kaggle Notebooks environment, leveraging dual NVIDIA Tesla T4 GPUs (2×16 GB VRAM) for accelerated computation. 
%During the training phase, input images were normalized using ImageNet statistics ($\mu$=[0.485,0.456,0.406],$\sigma$=[0.229,0.224,0.225]) and subjected to random horizontal and vertical flips (p=0.5). At inference time, the YOLO ensemble utilized built-in test-time augmentation (TTA) to improve detection robustness, while the U-Net architecture implemented multi-scale inference across a scale set of \{0.8, 1.0, 1.2\} to optimize localization accuracy across varying spatial resolutions.

\section{Results and Discussion}
\label{sec:results}
We evaluated the performance of our approach using TP, FP, Recall, Precision, F1 and mAP50-95. Table \ref{fig:table} reports the detection metrics for each model of our ensemble approach, as well as the final step of the optimization pipeline for the validation dataset.

The YOLOv8n 50×50 model achieved the highest sensitivity among the standalone detectors, yielding 2665 True Positives (TP) and a recall of 0.9937, at the cost of an excessive number of false detections (37398 FP). By ensembling the two YOLO variants, we significantly reduced the FP count while maintaining a stable TP rate. In contrast, while the U-Net model demonstrated the lowest TP count (2141), it produced a substantially higher F1-score (0.7425) due to its better precision and minimal FP rate compared to the YOLO-based architectures. Ultimately, the final ensemble, refined through the proposed optimization pipeline, achieved the peak mAP50-95 score of 0.6232, as shown in Table \ref{fig:table}.

While the final ensemble was optimized to maximize the mAP50-95 metric, this configuration resulted in an increased FP count relative to the standalone U-Net model. Although this setup was selected to satisfy the competition's primary goal of optimizing spatial localization accuracy, its practical clinical utility requires further consideration. In medical screening scenarios, a model that prioritizes a higher precision-to-recall ratio (F1) with fewer false alarms may be better suited for diagnostic assistance, even if it produces a marginal reduction in the mAP50-95 index.

\begin{figure}[t]
  \centering % Reemplaza \begin{center} para eliminar espacio extra
  \includegraphics[height=7.5cm, width=8.5cm]{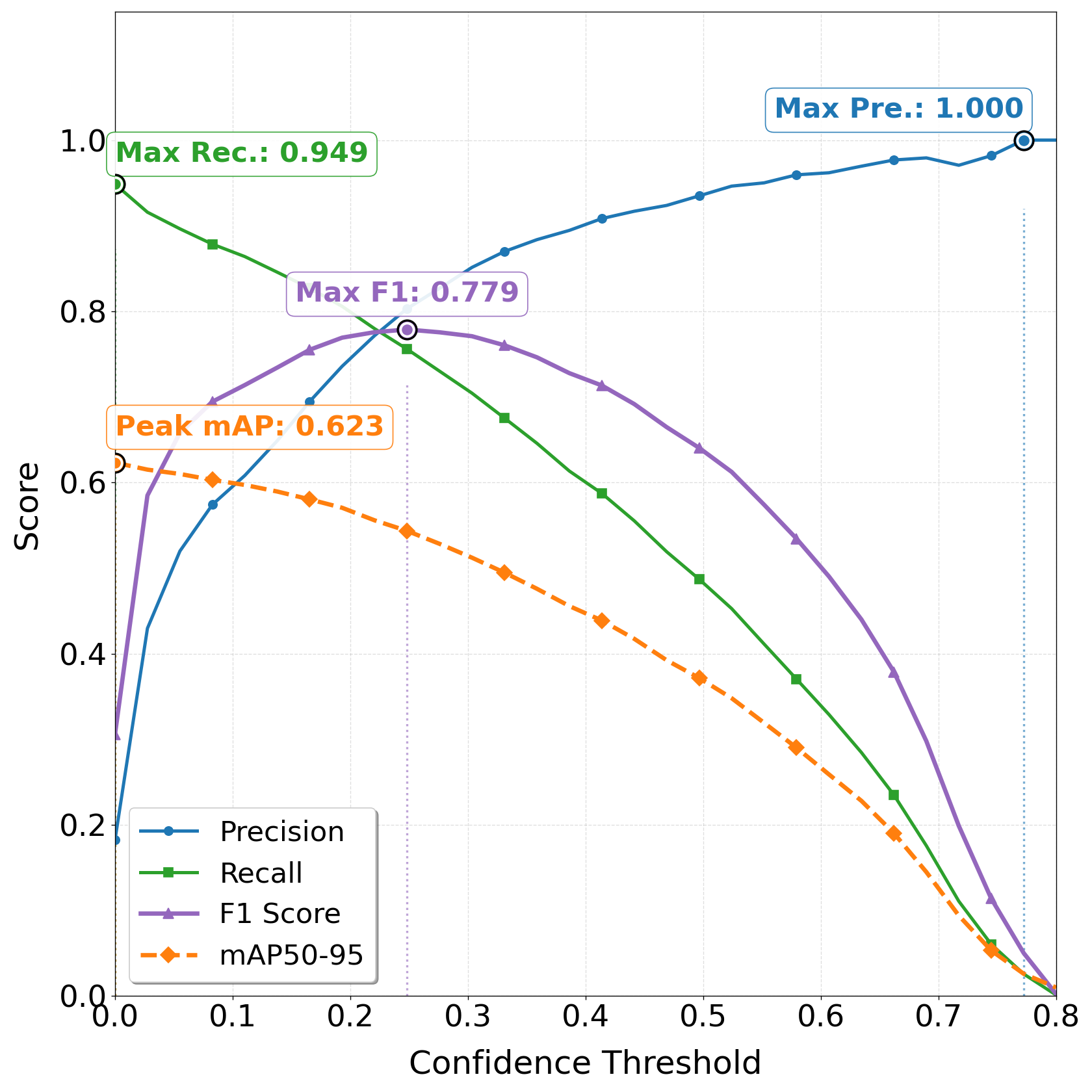}
  
  \vspace{-10pt} % Reduce el espacio entre la imagen y el caption
  \caption{Comprehensive performance metrics vs. confidence threshold for the proposed ensemble model. The graph highlights the peaks of each metric.}
  \label{fig:graph}
  \vspace{-10pt} % Reduce el espacio entre el caption y el texto de abajo
\end{figure}

Figure \ref{fig:graph} illustrates the performance metrics (Recall, Precision, F1-score, and mAP50-95) across varying confidence thresholds for the final prediction. As observed, optimizing for mAP50-95 requires adopting a lower confidence threshold, whereas other individual metrics peak at higher threshold values. Notably, the F1-score, which balances the inherent trade-off between precision and recall, reaches its maximum at a higher confidence level.

\section{Conclusions}
\label{sec:conclusion}

We presented a multi-model ensemble pipeline combining two YOLOv8 detectors and a U-Net heatmap regression model for automatic detection of cells in Pap smear images under the Bethesda classification system. The system achieves very high recall, reflecting the suitability of these architectures for this type of cytological detection task.

However, optimizing for mAP50-95 inherently encourages high recall at the cost of low-confidence false positives. We observe that modest increases in the confidence threshold yield substantial gains in precision with only marginal mAP reduction, suggesting the models produce well-calibrated scores. This trade-off is particularly relevant in clinical settings, where precision directly impacts patient management decisions.

\section{Acknowledgments}
\label{sec:acknowledgments}

We acknowledge Perez Bianchi et al. (2025) \cite{Perez_2025_RIVA} for making the data available.

This work was supported by Universidad de Buenos Aires.

\bibliographystyle{IEEEbib}
\bibliography{strings,refs}

\end{document}